\documentclass[conference]{IEEEtran}

\usepackage{cite}
\usepackage{amsmath,amssymb,amsfonts}
\usepackage{algorithmic}
\usepackage{graphicx}
\usepackage{textcomp}
\usepackage{xcolor}
\usepackage{hyperref}
\usepackage{tabularray}
\usepackage{bm}

\usepackage[switch]{lineno}

\def\BibTeX{{\rm B\kern-.05em{\sc i\kern-.025em b}\kern-.08em
    T\kern-.1667em\lower.7ex\hbox{E}\kern-.125emX}}
    
\begin{document}

\pagestyle{plain}
\renewcommand*{\thepage}{\small{\arabic{page}}}

\title{Reinforcement Learning for Efficient Toxicity Detection in Competitive Online Video Games}

\makeatletter
\newcommand{\linebreakand}{%
  \end{@IEEEauthorhalign}
  \hfill\mbox{}\par
  \mbox{}\hfill\begin{@IEEEauthorhalign}
}
\makeatother

\author{\IEEEauthorblockN{Jacob Morrier}
\IEEEauthorblockA{\textit{Division of the Humanities and Social Sciences} \\
\textit{California Institute of Technology} \\
Pasadena, CA, USA \\
\href{mailto:jmorrier@caltech.edu}{jmorrier@caltech.edu}}
\and
\IEEEauthorblockN{Rafal Kocielnik}
\IEEEauthorblockA{\textit{Computing + Mathematical Sciences} \\
\textit{California Institute of Technology} \\
Pasadena, CA, USA \\
\href{mailto:rafalko@caltech.edu}{rafalko@caltech.edu}}
\linebreakand
\IEEEauthorblockN{R. Michael Alvarez}
\IEEEauthorblockA{\textit{Division of the Humanities and Social Sciences} \\
\textit{California Institute of Technology} \\
Pasadena, CA, USA \\
\href{mailto:rma@caltech.edu}{rma@caltech.edu}}
}

\maketitle

\begin{abstract}
    Online platforms take proactive measures to detect and address undesirable behavior, aiming to focus these resource-intensive efforts where such behavior is most prevalent. This article considers the problem of efficient sampling for toxicity detection in competitive online video games. To make optimal monitoring decisions, video game service operators need estimates of the likelihood of toxic behavior. If no model is available for these predictions, one must be estimated in real time. To close this gap, we propose a contextual bandit algorithm that makes monitoring decisions based on a small set of variables that, according to domain expertise, are associated with toxic behavior. This algorithm balances exploration and exploitation to optimize long-term outcomes and is deliberately designed for easy deployment in production. Using data from the popular first-person action game \emph{Call of Duty\textregistered: Modern Warfare\textregistered III}, we show that our algorithm consistently outperforms baseline algorithms that rely solely on players' past behavior. This finding has substantive implications for the nature of toxicity. It also illustrates how domain expertise can be harnessed to help video game service operators identify and mitigate toxicity, ultimately fostering a safer and more enjoyable gaming experience.
\end{abstract}


\section{Introduction}

\IEEEPARstart{T}{he} detrimental effects of toxicity in online platforms are widely recognized and well-documented \cite{CAPLAN20091312,  gray_2012, salter_blodgett_2012, kuznekoff_rose_2013, FOX2014314, chess_shaw_2015, kwak_blackburn_han_2015, ballard_2017, madden_2021, turkay_2020, kowert_kilmer, zsila2022toxic, morrier2024uncoveringeffecttoxicityplayer}. This prompts video game service operators to adopt proactive measures to monitor and address such behavior. However, these efforts are resource-intensive. Consequently, operators seek to focus them where toxicity is most prevalent, ensuring their interventions have the most impact.

This article considers the problem of efficient sampling for toxicity detection in competitive online video games. At the beginning of each match, video game service operators must choose whether to monitor each player's in-game voice interactions. In doing so, their objective is to maximize the detection of toxic behavior while minimizing monitoring costs, as measured by the volume of monitored voice interactions. To achieve this, service operators wish to monitor only instances where the likelihood of toxic behavior exceeds some threshold.

To make informed monitoring decisions, video game service operators need estimates of the likelihood of toxic behavior. However, a predictive model may not always be available, especially immediately after a game's release. In such cases, operators must estimate a model, a process that inherently involves exploration. This implies that they may find it valuable to monitor players' in-game voice interactions not only when they are confident that toxic behavior is likely, making optimal monitoring decisions based on existing data, but also when uncertainty is high, even if toxicity seems a priori unlikely, to gather more data and refine future predictions. Throughout this process, the overarching objective of video game service operators is to balance exploration and exploitation to optimize long-term outcomes.

To resolve the tension between static and dynamic incentives, we propose a reinforcement learning algorithm that adaptively learns where computational resources for toxicity detection should be optimally allocated and makes monitoring decisions accordingly. This contextual bandit algorithm chooses which players' in-game voice interactions to monitor to maximize the detection of toxic behavior while minimizing monitoring costs, that is, by monitoring the fewest players in the fewest matches necessary. It bases its decisions on a limited number of readily observable variables that are, according to domain expertise, associated with toxic behavior. This algorithm was deliberately designed for easy deployment in production.

To evaluate our proposed algorithm's effectiveness, we compare its performance with that of two baseline algorithms making decisions based on individual players' past behavior or, in other words, on whether they have previously engaged in such behavior. We perform this analysis within the context of the popular first-person action video game \emph{Call of Duty\textregistered: Modern Warfare\textregistered III}, focusing specifically on its primary multiplayer game mode, Team Deathmatch. The results show that our algorithm consistently outperforms the alternatives. This finding suggests that some non-idiosyncratic contextual factors are strongly associated with a higher likelihood of players engaging in toxic behavior, allowing for the effective optimization of monitoring strategies. Remarkably, decisions based on these factors result in systematically better outcomes than those based solely on individual players' past tendencies toward toxicity, challenging the perspective that toxicity is an inherently idiosyncratic phenomenon.

This article not only corroborates the correlation between contextual features and toxic behavior and compares it to the predictive power of individual players' past behavior for future toxic behavior, thereby providing insights into the nature of toxicity. It also harnesses these findings to improve the efficiency of toxicity detection efforts through advanced computational methods. Accordingly, this research supports and strengthens video game service operators' efforts to identify and mitigate toxicity in competitive online video games, ultimately fostering a safer and more enjoyable gaming experience.

This article is structured as follows. We begin by discussing how our work relates to the previous literature. Next, we describe the data used in our analysis. We then define the optimization problem faced by video game service operators, review factors known to be correlated with toxicity, and outline our proposed algorithm. Afterward, we present the results of an experiment that compares the performance of our algorithm with two baseline algorithms. We conclude by discussing the implications of our findings and outlining potential avenues for future research.

\section{Related Literature}

This article primarily contributes to the burgeoning literature on reinforcement learning for content moderation on online platforms \cite{avadhanula2022banditsonlinecalibrationapplication}. Our contribution is applied, demonstrating how video game service operators, drawing on their domain expertise, can implement contextual bandit algorithms to optimize their content moderation efforts. Our study also relates to the extensive literature on content moderation and the detection of undesirable behavior in online communities, presenting an approach for enhancing the efficiency of these computationally intensive efforts \cite{Chen_2012, Martens_2015, Nobata_2016, Zhong_2016, malmasi-zampieri-2017-detecting, Van_Hee_2018, Sanzgiri_2018, Cambridge_2019, Gampa_2023_CVPR}. Lastly, we draw on insights from the literature on toxicity in competitive online video games to identify contextual features that can predict toxic behavior and, by extension, inform monitoring decisions \cite{FOX2014314,grandprey2014identification,ijerph17176429,kordyaka_2020,Kou_2020,mclean2020toxic,SHEN2020106343,beres2021don,zsila2022toxic,kocielnik2024challenges,kumar2024emotionalimpactgameduration,morrier2024uncoveringeffecttoxicityplayer}.

\section{Data}

We analyze proprietary data from \emph{Call of Duty}, a popular first-person action video game franchise published by Activision\textregistered. We focus on \emph{Call of Duty: Modern Warfare III}, and especially its most popular multiplayer game mode, Team Deathmatch. In this mode, players are divided into two equally sized teams and compete to achieve the highest number of eliminations. After a brief pause, eliminated players reappear at a different location on the map. A team wins by reaching a predetermined elimination limit first or accumulating the most eliminations by the end of the match.

Since 2023, Activision has collaborated with Modulate\texttrademark, a startup developing intelligent voice technology to identify and combat online toxicity, and integrated its proprietary voice chat moderation technology, ToxMod\texttrademark, into its gaming platforms \cite{toxmod_description}. ToxMod is a voice moderation technology that analyzes online speech for emotion, volume, transcribed content and intention, and other related signals to identify harmful or malicious content \cite{kowert_woodwell}. These signals are fed into machine learning models that classify the primary type of harm in an audio clip. The voice chat moderation technology's initial beta rollout began in North America on August 30, 2023, within \emph{Call of Duty: Modern Warfare II} and \emph{Call of Duty: Warzone\texttrademark}. This was followed by a global release (excluding the Asia-Pacific region) that coincided with the launch of \emph{Call of Duty: Modern Warfare III} on November 10, 2023. ToxMod only supported English during our observation period.

ToxMod offers extraordinary data on toxic behavior in competitive online video games, serving as the basis of our analysis. Our data set includes data from a sample of matches in Team Deathmatch mode monitored by ToxMod during the first month following the game's launch, covering the period from November 10 to December 10, 2023. Overall, our sample consists of 207,338,296 observations, each representing a player in a match, drawn from 15,644,547 matches and 8,798,876 players. A player is classified as having engaged in toxic behavior during a game if ToxMod flagged at least one of their voice interactions during that match as toxic.

\section{Problem Formulation}

We provide a formal definition of video game service operators' decision-making problem regarding whether to monitor a player's in-game voice interactions. We consider a setting in which service operators can choose, at the start of every match, to monitor a player's in-game voice interactions for the duration of that match. Operators seek to monitor in-game voice interactions to detect and eventually address toxic behavior. However, monitoring is costly and provides no actionable insight when no toxicity is detected. In this context, operators wish to maximize the detection of offenses while minimizing the volume of monitored in-game voice interactions.

\begin{table}
    \centering
    \caption{Reward Structure}
    \label{tab:rewards}
    \begin{tblr}{colspec = {XXXX}, columns = {halign = c, valign = m}, rows = {rowsep = 5pt}, hline{1-2}={3-5}{solid}, hline{3-5}, vline{1-2}={3-5}{solid}, vline{3-5}}
         & & \SetCell[c=2]{c} \textbf{Player Behavior} & \\
         & & \textit{Toxicity} & \textit{No Toxicity} \\
         \SetCell[r=2]{c} \textbf{Monitoring Decision} & \textit{Monitor} & $1 - c$ & $-c$ \\
         & \textit{Not Monitor} & \SetCell[c=2]{c} 0 & \\
    \end{tblr}
\end{table}

Formally, if the service operator opts to monitor a player's in-game voice interactions, we assume that they incur a fixed cost $c$. In turn, the operator earns rewards that depend on the player's behavior: they receive a reward of 1 if toxic behavior is detected and a reward of 0 otherwise. On the other hand, if the operator chooses not to monitor the player's voice interactions, they receive a fixed reward of 0 regardless of the player's behavior. The rewards conditional on the monitoring decision and the player's behavior are summarized in Table \ref{tab:rewards}.

In this framework, service operators wish to selectively monitor players' in-game voice interactions based on their behavior. Specifically, they prefer to monitor a player's in-game voice interactions when they engage in toxic behavior and not to monitor them when they do not engage in such behavior. Given the uncertainty of the player's behavior, these preferences translate into the following static decision rule: it is optimal to monitor a player's in-game voice interactions when the likelihood of toxic behavior surpasses the cost of monitoring:
\begin{equation*}
    P\left(\text{Player engages in toxic behavior}\right) > c.
\end{equation*}

In other words, to optimize resource usage, service operators should monitor a player's in-game voice interactions only when the likelihood of toxic behavior is sufficiently high, ensuring that the gathered data is actionable.

To implement this static decision rule, video game service operators need estimates of the likelihood of toxic behavior. However, a pre-existing model may not always be available for making these predictions, such as after a game's release. In this scenario, operators must estimate a model in real time. This introduces a second dynamic decision-making problem: operators wish to monitor players' in-game voice interactions not only when they are confident that toxic behavior is likely, in which case they \emph{exploit} the available data, but also when uncertainty is high, in which case they \emph{explore} to refine predictions and improve future decisions. As operators explore and collect more data, they seek to optimize long-term outcomes from a static perspective by exploiting as much information as possible. This exploration-exploitation trade-off is the target of bandit algorithms.

\section{Correlates of Toxicity}

\begin{table}
    \centering
    \caption{Toxicity Correlates}
    \label{tab:correlates}
    \begin{tblr}{colspec={XXX}, columns={halign=l, valign=m}, column{3}={halign=c}, row{1}={font=\bfseries, halign=c}, hlines, vlines}
        Variable & Expected Relationship with Toxicity & References \\
        Skill Level & Experienced and highly skilled players may exhibit a greater tendency toward toxic behavior. & \cite{grandprey2014identification,kordyaka_2020,SHEN2020106343} \\
        Average Skill Difference with Opponents & Players often display toxic behavior toward those they view as ``outsiders,'' such as lower-skilled players.  & \cite{FOX2014314,Kou_2020,SHEN2020106343} \\
        Average Skill Difference with Teammates & Heightened competitiveness can contribute to increased toxicity. & \cite{Kou_2020,SHEN2020106343} \\
        Presence of Teammates from the Same Party & \SetCell[r=2]{m} Toxic behavior may occur more frequently in the presence of familiar individuals, such as teammates from the same party. & \SetCell[r=2]{m} \cite{grandprey2014identification,kordyaka_2020,Kou_2020,mclean2020toxic,SHEN2020106343,zsila2022toxic,morrier2024uncoveringeffecttoxicityplayer} \\
        Proportion of Teammates from the Same Party & & \\
        Matches Played in the Current Session & Players may be more likely to exhibit toxic behavior after playing many matches in a single session. & \cite{ijerph17176429,kumar2024emotionalimpactgameduration} \\
        Reports Filed Against the Player in the Last 24 Hours & Reports against players may reflect past toxic behavior. & \cite{grandprey2014identification,beres2021don,zsila2022toxic,kocielnik2024challenges} \\
        Reports Filed by the Player in the Last 24 Hours & Players exposed to toxic behavior are more prone to engage in similar behavior, and filed reports may serve as markers of prior exposure to toxicity. & \cite{FOX2014314,kordyaka_2020,kocielnik2024challenges} \\
    \end{tblr}

    \begin{flushleft}
        \emph{Note:} A party consists of many players voluntarily playing together as a single, cohesive unit.
    \end{flushleft}
\end{table}

\begin{table}
    \centering
    \caption{Descriptive Statistics}
    \label{tab:statistics}
    \begin{tblr}{colspec={Xccccc}, columns={valign = m}, column{1}={halign = l}, row{1}={font=\bfseries, halign = c}, hlines, vlines}
        Variable & Mean & Std. Error & Min. & Median & Max. \\
        Toxic Behavior & 0.000372 & 0.0193 & 0 & 0 & 1 \\
        Skill Level & $-$43.994 & 207.828 & $-$736 & $-$45 & 716 \\ 
        Average Skill Difference with Opponents & 96.025 & 76.750 & 0 & 75 & 1,061.8 \\
        Average Skill Difference with Teammates & 103.586 & 85.075 & 0 & 78.833 & 1,022.8 \\
        Presence of Teammates from the Same Party & 0.336 & 0.472 & 0 & 0 & 1 \\
        Proportion of Teammates from the Same Party & 0.104 & 0.185 & 0 & 0 & 1 \\
        Matches Played in the Current Session & 3.751 & 5.840 & 0 & 2 & 150 \\
        Reports Filed Against the Player in the Last 24 Hours & 0.0364 & 0.236 & 0 & 0 & 153 \\
        Reports Filed by the Player in the Last 24 Hours & 0.0449 & 0.914 & 0 & 0 & 304 \\
    \end{tblr}
\end{table}

\begin{table}
    \centering
    \caption{Regression Results}
    \label{tab:regression}
    \begin{tblr}{colspec={Xcc}, column{1}={halign=l}, rowsep=0pt, row{odd}={abovesep=2pt}, row{even}={belowsep=2pt}}
    \hline
    \hline
    & (1) & (2) \\
    & Linear$^{\dag}$ & Logistic \\
    \hline
     \SetCell[r=2]{t} Skill Level & 0.046$^{***}$ & 0.001$^{***}$ \\
    & (0.001) & (0.000) \\
     \SetCell[r=2]{t} Average Skill Difference with Opponents & -0.036$^{***}$ & -0.001$^{***}$ \\
    & (0.005) & (0.000) \\
     \SetCell[r=2]{t} Average Skill Difference with Teammates & -0.004$^{}$ & -0.000$^{*}$ \\
    & (0.005) & (0.000) \\
     \SetCell[r=2]{t} Presence of Teammates from the Same Party & 48.603$^{***}$ & 1.566$^{***}$ \\
    & (0.656) & (0.011) \\
     \SetCell[r=2]{t} Proportion of Teammates from the Same Party & 66.239$^{***}$ & 0.697$^{***}$ \\
    & (1.928) & (0.020) \\
     \SetCell[r=2]{t} Matches Played in the Current Session & -0.145$^{***}$ & -0.001$^{}$ \\
    & (0.024) & (0.001) \\
     \SetCell[r=2]{t} Reports Filed Against the Player in the Last 24 Hours & 62.794$^{***}$ & 0.251$^{***}$ \\
    & (1.520) & (0.004) \\
     \SetCell[r=2]{t} Reports Filed by the Player in the Last 24 Hours & 6.134$^{***}$ & 0.033$^{***}$ \\
    & (0.299) & (0.001) \\
    \hline
    \hline
    \SetCell[c=3]{l} \textit{Note:} \hspace{2.05cm}$^{\dag}\times 10^{-6}$; $^{*}p<0.1$; $^{**}p<0.05$; $^{***}p<0.01$ \\
    \end{tblr}
\end{table}

Past research on toxicity in competitive online video games has identified contextual features correlated with such behavior. Many of these features are easily observable before the beginning of a match, making them suitable for predicting the likelihood of toxic behavior. This prediction can then be used to steer monitoring decisions.

Table \ref{tab:correlates} lists eight of these variables, describing their expected relationship with toxicity and supported by relevant references. Descriptive statistics for these variables and toxic behavior in our data set are presented in Table \ref{tab:statistics}. Table~\ref{tab:regression} contains the estimates of linear and logistic regressions of a player's behavior during a match with these covariates. The linear regression coefficient and standard error estimates are all scaled by a $10^{-6}$ factor.

The regression results indicate that all covariates, except for the average skill difference with teammates and the number of matches played in the current session, consistently exhibit a statistically significant relationship with toxic behavior. This implies that these variables can predict toxic behavior and, in turn, guide monitoring decisions.

However, this conclusion is drawn a posteriori from all the observations in our data. Instead, we aim to assess whether these correlations can be learned and be effectively used to optimize monitoring decisions in real time over our period of interest.

\section{Bandit Algorithms}

\begin{figure*}
    \centering
    \includegraphics[width=\linewidth,page=1]{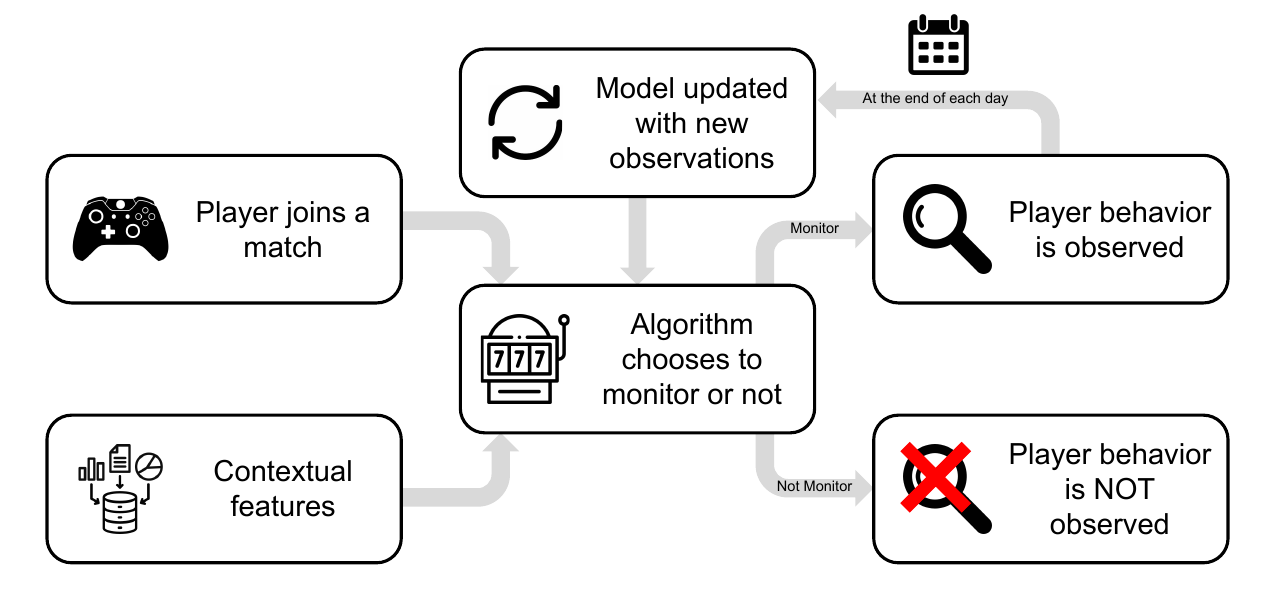}
    \caption{Algorithm Workflow}
    \label{fig:overview}
\end{figure*}

There is an evident analogy between the optimization problem defined above and bandit problems: at the start of every match and for every player, video game service operators must choose whether to pull the ``monitor'' arm or the ``not monitor'' arm. The ``monitor'' arm generates stochastic rewards determined by the player's behavior, whereas the ``not monitor'' arm generates fixed, known rewards.

We propose to make monitoring decisions using the LinUCB algorithm, with the variables listed above as input covariates \cite{LinUCB}. This algorithm assumes that the expected reward from monitoring a player's in-game voice interactions or, put differently, the likelihood that a player engages in toxic behavior is a linear function of the covariates, with an unknown coefficient vector estimated through ridge regression:
\begin{equation*}
    \Hat{\bm{\theta}} = \left(\bm{X}'\bm{X} + \bm{I}\right)^{-1} \bm{X}'\bm{y},
\end{equation*}
where $\bm{X}$ is a matrix containing the covariate values for all previously monitored matches and players, and $\bm{y}$ is a vector denoting whether toxic behavior was observed in these observations.

Given the resulting estimate of expected rewards, we apply an Upper Confidence Bound (UCB) arm-selection strategy: we pull the ``monitor'' arm if and only if the expected rewards plus the product of this expectation's standard error and a predefined exploration factor $\delta$ exceeds the monitoring cost $c$:
\begin{equation*}
    \underbrace{\bm{x}'\Hat{\bm{\theta}}}_{\text{\scriptsize \begin{tabular}{c}
        Expected \\
        Rewards
    \end{tabular}}} + \delta \underbrace{\sqrt{\bm{x}'\left(\bm{X}'\bm{X} + \bm{I}\right)^{-1}\bm{x}}}_{\text{Standard Error}} > c,
\end{equation*}
where $\bm{x}$ is an observation's covariate vector.

The exploration factor reflects the weight service operators assign to gathering more information to improve future decisions. A higher value of this parameter implies a greater inclination for exploration. On the other hand, the cost parameter defines how much operators wish to confine monitoring to contexts with a high likelihood of toxic behavior.

At the end of each day, the model parameters are updated using the observations gathered throughout the day. The updated model is then used to make monitoring decisions for the following day. Unlike updating the model after every observation, daily updates allow for batch processing, lowering computational costs significantly, while still enabling continuous learning. The workflow of the algorithm is illustrated in Fig. \ref{fig:overview}.

This bandit algorithm addresses the ``cold-start problem'' faced by video game service operators after a game's release. Since there is no pre-existing model for predicting toxic behavior and informing monitoring decisions, one must be estimated in real time. This inherently involves some exploration. In this context, monitoring a player's voice interactions can still be valuable even when the immediate cost exceeds the expected rewards because the collected data improves future predictions and decisions. Over time, as more data is gathered, the algorithm gradually shifts toward exploitation, relying primarily on expected rewards to make monitoring decisions. However, when the algorithm lacks enough information to make an informed choice, it retains the option to keep exploring. Also, by continuously updating its parameters, the algorithm can adapt to evolving conditions.

We compare the performance of our proposed algorithm against two baseline algorithms: the deterministic and probabilistic Explore-Then-Commit algorithms \cite{Lattimore_Szepesvari_2020}. Unlike LinUCB, which optimizes decisions by pooling information across all players, these algorithms identify optimal monitoring decisions based on individual players' prior history of toxic behavior. This is based on the reasonable premise that past toxic players are more likely to engage in similar behavior in the future. The deterministic Explore-Then-Commit algorithm involves monitoring each player for a fixed and predetermined number of matches. Monitoring continues beyond this exploratory period if the player is observed engaging in toxic behavior at least once. In contrast, the probabilistic Explore-Then-Commit algorithm consists of randomly monitoring a fixed share of observations, and monitoring continues for a player if they are observed to engage in toxic behavior at least once. This algorithm is equivalent to a $\varepsilon$-greedy algorithm in which operators explore by monitoring with a fixed probability and choose whether to monitor or not based on past behavior with complementary probability.

Initially, all matches and players in our dataset are presumed to be monitored. However, we conduct an experiment in which not all matches and players are monitored and monitoring decisions are made by the bandit algorithms described earlier. In doing so, our goal is to assess the performance of these algorithms in choosing which players and matches to monitor. To reflect the objectives of video game service operators, we evaluate the performance of the algorithms based on detection rate or recall, defined as the share of toxic behavior detected relative to all toxic behavior that occurred. For a given proportion of matches and players monitored, which reflects the level of monitoring cost, the optimal algorithm is the one that maximizes the detection rate.

\section{Results}

\begin{figure}
    \centering
    \includegraphics[width=\linewidth]{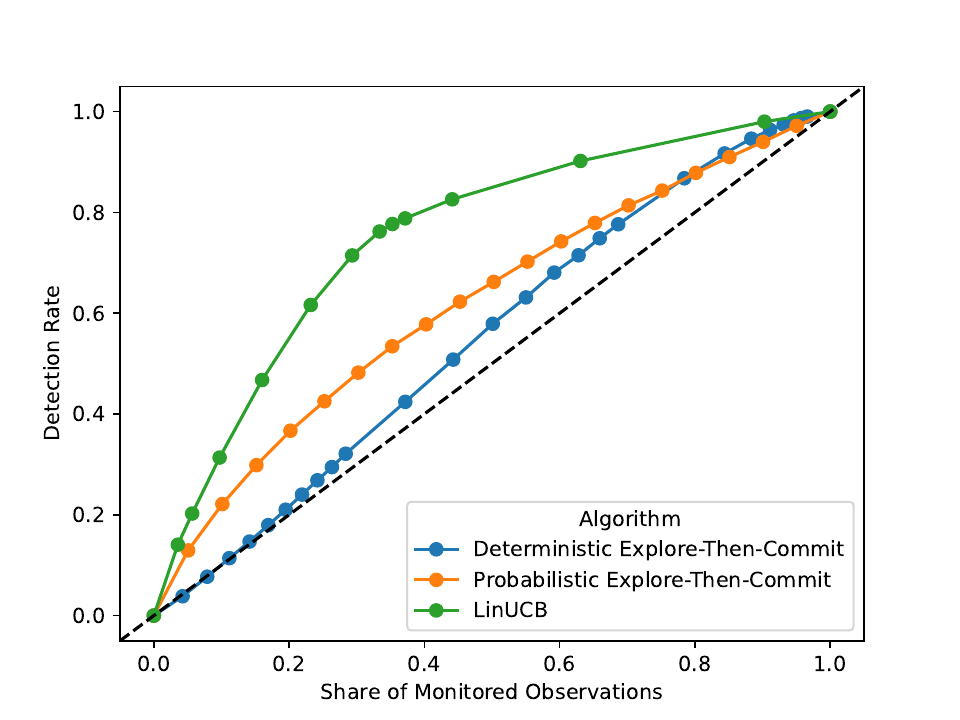}
    \caption{Performance Comparison of Bandit Algorithms}
    \label{fig:performance}
\end{figure}

\begin{table*}
    \centering
    \caption{Performance Comparison of the LinUCB and Probabilistic Explore-Then-Commit Algorithms}
    \label{tab:performance}
    \begin{tblr}{colspec={XXccccccccc}, columns={halign=c, valign=m}, column{1}={font=\bfseries}, row{1}={font=\bfseries}, vlines, hlines}
        \SetCell[c=2]{c} Share of Monitored Observations & & 0.1 & 0.2 & 0.3 & 0.4 & 0.5 & 0.6 & 0.7 & 0.8 & 0.9 \\
        \SetCell[r=2]{m} Detection Rate & \emph{Probabilistic Explore-Then-Commit} & 0.2196 & 0.3634 & 0.4769 & 0.5751 & 0.6604 & 0.7383 & 0.8112 & 0.8779 & 0.9405 \\
        & \textit{LinUCB} & 0.3202 & 0.55 & 0.7225 & 0.8035 & 0.8496 & 0.8897 & 0.9219 & 0.9506 & 0.9792 \\
        \SetCell[r=2]{m} \textbf{Improvement} & \textit{pp.} & 10.06 & 18.66 & 24.56 & 22.84 & 18.92 & 15.14 & 11.07 & 7.27 & 3.87 \\
        & \textit{\%} & 45.81 & 51.35 & 51.5 & 39.71 & 28.65 & 20.51 & 13.65 & 8.28 & 4.11 \\
    \end{tblr}
\end{table*}

Fig. \ref{fig:performance} depicts the detection rate of toxic behavior as a function of the proportion of monitored observations for each algorithm. For any given share of monitored observations, our proposed LinUCB algorithm consistently outperforms the baseline algorithms in detection rate. The probabilistic Explore-Then-Commit algorithm ranks second, except for the highest share of monitored observations.

Table \ref{tab:performance} presents a detailed performance comparison of the proposed LinUCB and probabilistic Explore-Then-Commit algorithms. Again, we see that LinUCB exhibits a consistently superior performance, achieving an increase in detection rate of up to 24.56 percentage points (pp.), equivalent to a 51.5\% improvement. In other words, with constant monitoring costs, LinUCB can detect up to 51.5\% more offenses, providing an equivalent volume of actionable insights, than the probabilistic Explore-Then-Commit algorithm.

\section{Discussion and Conclusion}

In this paper, we considered the problem of efficient sampling for toxicity detection in competitive online video games. Using data from the popular first-person action video game \emph{Call of Duty: Modern Warfare III}, we evaluated the performance of various bandit algorithms in optimizing monitoring decisions. Our proposed LinUCB algorithm with a small set of contextual features exhibits a superior performance over random sampling and two baseline algorithms by detecting a higher share of offenses for a given volume of monitored in-game voice interactions. Consequently, our LinUCB algorithm has the potential to support and strengthen video game service operators' efforts to detect and mitigate toxicity in competitive online video games by making them more efficient, ultimately fostering a safer and more enjoyable gaming environment.

The superior performance of the LinUCB algorithm over benchmark algorithms has substantive implications for the nature of toxicity. One perspective views toxicity as an inherently idiosyncratic and individual phenomenon, largely independent of context, with a small group of players recurrently and spontaneously engaging in such behavior. An alternative perspective considers toxicity as context-dependent, influenced by specific contextual features influencing a player's likelihood of engaging in toxic behavior. Our findings challenge the first perspective by showing that: (i) a few contextual features are strongly associated with an increased likelihood of toxic behavior, and (ii) monitoring decisions based on these factors are more effective than those based on players' past tendencies toward toxicity.

In conclusion, multiple avenues are available for expanding on this research. For instance, future research should explore additional covariates to improve the accuracy of toxic behavior predictions, including accounting for past moderation actions. Evaluating our proposed algorithm's performance beyond a one-month period and in other video games would also be valuable. Lastly, investigating the application of bandit algorithms to enhance and optimize human moderation efforts, arguably even more costly than automated toxicity detection, offers a compelling direction for further research.

\section*{Acknowledgment}

The authors express their gratitude to Andrea Boonyarungsrit, Grant Cahill, MJ Kim, Jonathan Lane, Amine Mahmassani, Myrl Marmarelis, Gary Quan, Deshawn Sambrano, Feri Soltani, and Michael Vance for their invaluable feedback and support in writing this article. The views and opinions presented are solely those of the authors and do not reflect those of Activision\textregistered.

\bibliographystyle{IEEEtran}
\bibliography{bibliography}

\end{document}